\documentclass{article}
\usepackage{ijcai19}

\usepackage{times}
\usepackage{color}
\usepackage{soul}
\usepackage{url}
\usepackage[hidelinks]{hyperref}
\usepackage[utf8]{inputenc}
\usepackage[small]{caption}
\usepackage{graphicx}
\usepackage{amsmath}
\usepackage{booktabs}
\usepackage{makecell} 
\usepackage{arydshln} 
\usepackage{multirow}
\usepackage{subfig}
\newcommand{\argmin}{\mathop{\rm argmin}\limits}

\title{Learning Fine Grained Place Embeddings with Spatial Hierarchy\\ from Human Mobility Trajectories}

\author{
Toru Shimizu$^1$\footnote{Contact Author}\and
Takahiro Yabe$^2$\and
Kota Tsubouchi$^1$\\
\affiliations
$^1$Yahoo Japan Corporation, Japan\\
$^2$Lyles School of Civil Engineering, Purdue University, USA\\
\emails
\{toshimiz, ktsubouc\}@yahoo-corp.jp,
tyabe@purdue.edu
}

\begin{document}

\maketitle

\begin{abstract}
Place embeddings generated from human mobility trajectories have become a popular method to understand the functionality of places. 
Place embeddings with high spatial resolution are desirable for many applications, however, downscaling the spatial resolution deteriorates the quality of embeddings due to data sparsity, especially in less populated areas. 
We address this issue by proposing a method that generates fine grained place embeddings, which leverages spatial hierarchical information according to the local density of observed data points. 
The effectiveness of our fine grained place embeddings are compared to baseline methods via next place prediction tasks using real world trajectory data from 3 cities in Japan. 
In addition, we demonstrate the value of our fine grained place embeddings for land use classification applications. 
We believe that our technique of incorporating spatial hierarchical information can complement and reinforce various place embedding generating methods.

\end{abstract}

\section{Introduction}

\begin{figure}[t]
  \includegraphics[width=1.0\linewidth]{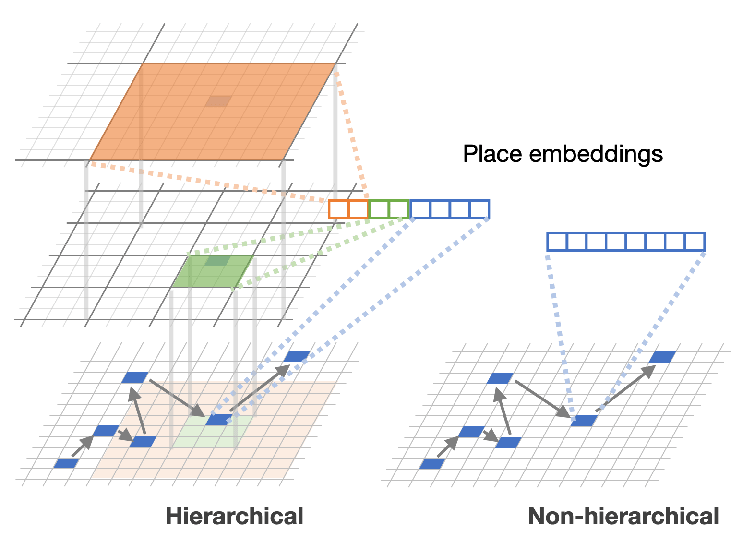}
  \caption{Hierarchical and non-hierarchical place embeddings learned by next place prediction.}
  \label{fig:hier_embd_overview}
\end{figure}

Generating embeddings of places from human mobility trajectories via neural network models has become a popular technique for understanding the functionality of urban places.
Such place embeddings have been shown to be effective in solving various urban challenges including land use classification (e.g. \cite{yao2018representing}) and place-of-interest recommendation (e.g. \cite{chang2018content}). 
A recent study applied neural machine translation techniques to investigate the transferability of place embeddings across different cities, at a 1km spatial scale \cite{city2city}. 
In many of these tasks, place embeddings with finer spatial resolution would be desirable, since they would be able to provide details about the functionality of specific locations. 

However, downscaling the spatial resolution deteriorates the quality of embeddings due to sparsity of data used to train the model. 
Although we are able to learn place embeddings with high spatial resolution in urban areas with large amount of human mobility flow, this downscaling of place embeddings often fails in rural areas where data are sparse. 

In this study, we propose a novel method to generate fine grained place embeddings that overcome the aforementioned challenge.
To achieve such improvements, we leverage spatial hierarchical information when generating the place embeddings. 
Our method employs different strategies for embedding generation depending on the availability of input data.
More specifically, embeddings are generated based on the locally observed mobility trajectories in urban areas with plenty of data, but for rural areas with less data, embeddings are generated by complementing information from surrounding places which are in the upstream of the spatial hierarchy. 


In summary, the contributions of this study are as follows:
\begin{itemize}
    \item We propose a novel method that integrates spatial hierarchical information to improve the quality of generated place embeddings from mobility data.
    \item We show the effectiveness of our hierarchical place embeddings compared to baseline methods via next place prediction tasks using real world trajectory data from 3 cities in Japan. 
    \item We demonstrate the value of our hierarchical place embeddings in land use classification applications.
\end{itemize}

\section{Data}
\textbf{Smartphone GPS Data:}
Yahoo Japan Corporation\footnote{\url{https://about.yahoo.co.jp/info/en/company/}} collects location information of mobile phone app users in order to send relevant notifications and information to the users. 
In this study, we use trajectory data collected from a total of more than 1 million mobile phone users estimated to be living in the 3 target cities (Kyoto, Fukuoka, Okayama). 
The users in this study have accepted to provide their location information. 
The data are anonymized so that individuals cannot be specified, and personal information such as gender, age and occupation are unknown. 
Each GPS record consists of a user's unique ID (random character string), timestamp, longitude, and latitude. 
The data acquisition frequency of GPS locations changes according to the movement speed of the user to minimize the burden on the user's smartphone battery. 
When the user is staying in a certain place for a long time, data is acquired at a relatively low frequency, and if it is determined that the user is moving, the data is acquired more frequently. 
The data has a sample rate of approximately 2\% of the population, and past studies suggest that this sample rate is enough to grasp the macroscopic urban dynamics. 

\noindent
\textbf{Urban Land Use Data:}
To validate whether the generated fine grained place embeddings correctly capture the functionality of the places, we use the Urban Area Land Use Mesh Data\footnote{\url{http://nlftp.mlit.go.jp/ksj/gml/datalist/KsjTmplt-L03-b-u.html}} in the National Land Numerical Information Database provided by the Ministry of Infrastructure, Land, and Transport and Tourism of Japan. 
The dataset divides all urban areas of the entire country into $100m \times 100m$ grid cells, and assigns one category to each grid cell out of 17 options. 
The 17 options include farmland, residential area, business district, parks, forests, factories, public facilities, water body, open spaces, roads, railways, golf courses, etc, shown in Table \ref{tbl:land_use_labels}. 
We aggregate these data into a larger spatial scale ($500m \times 500m$), thus for each place, we have a 17 dimensional vector where each element shows how many pixels of a specific land type exists in that place. 

\section{Preliminaries} 
First, we give brief explanations of key concepts that are used in this paper.

\noindent
\newline
\textit{Definition 1} (\textbf{Human Mobility Trajectories})
Human mobility trajectories are sequences of staypoint locations generated using location data collected from mobile phones. 
For each user, staypoints are detected using spatio-temporal clustering \cite{zheng2015trajectory}.
Each staypoint is then converted into place IDs, which are $125m \times 125m$ at the finest spatial scale. 
Human mobility trajectories also contain information on the timestamp of entry and exit from each staypoint, and the duration of stay at each staypoint. 

\noindent
\newline
\textit{Definition 2} (\textbf{Places}) Geographical spaces are divided into disjoint cells, referred to as ``places'' in this paper, by grid sizes of $r$ meters. 
Grid sizes $r$ of several spatial scales (10km, 1km, 500m) are used to integrate spatial hierarchical information when learning the embeddings of places.

\noindent
\newline
\textit{Definition 3} (\textbf{Place Embeddings})
We denote the embedding of each place $i$ as $\textbf{x}_i$, which is a $d$-dimensional vector.
Similar to word embeddings \cite{mikolov2013distributed}, place embeddings are vector representations of the functionality of the place learned from human mobility trajectories.
The place embeddings are generated via solving a next place prediction task using an LSTM RNN model, using human mobility trajectories as input data.

\section{Spatial Hierarchical Embeddings}

\subsection{Place Embeddings}
A trajectory $s$ in Human Mobility Trajectories $T$ contains a sequence of places and other attributes such as time stamps and duration of stays.
A trajectory $s \in T$ with the length $n$ is thus expressed as
\begin{align}
    s = \{m_1, \mathbf{a_1}, m_2, \mathbf{a_2}, \ldots, m_n, \mathbf{a_n}\}
\end{align}
where $m_t$ is a place and $\mathbf{a_t}$ is a feature vector representing other attributes at a time step $t$.
We can generate place embeddings such as $\mathbf{x}_{m_1}, \ldots, \mathbf{x}_{m_n}$ by recurrent neural networks (RNNs) with long short-term memory (LSTM) or gated recurrent units (GRU), as a sub-product of the task of predicting the next place a user will visit in the next time step given the previous information.
Specifically, the model parameters $\Theta$ including the place embeddings can be optimized while minimizing the cross entropy loss of predicting the next place
\begin{align}
    \hat{\Theta} &= 
    \argmin_{\Theta} \sum_{s \in T_{\textrm{train}}} \sum_{t=1}^{n_s-1} \notag \\
    & \qquad \big(
      -\log p_{\Theta} (m_t | m_1, \mathbf{a}_1, \ldots, m_{t-1}, \mathbf{a}_{t-1})
    \big)
\end{align}
where $\hat{\Theta}$ is the optimized set of parameters, $n_s$ is the sequence length of $s$, and $T_{\textrm{train}}$ is the training set of the input human mobility trajectories.
After that, a set of place embeddings corresponding to places seen in $T_{\textrm{train}}$, which we denote by $V$,
\begin{align}
    \{ \mathbf{x}_i | i \in V \}
\end{align}
can be obtained from the model and utilized for other tasks.

Also, in these kinds of models, the matrix in the output layer is similar in shape to the embedding matrix given the assumption that the same set of places $V$ are used for both input and output. 
This characteristic makes their parameters shareable.
In this work, we assume that places are represented by fine grained grid cells and that places are visited in a long-tail manner, i.e., the amount of trajectories' steps per place is sparse in the tail places.
As a natural consequence, we assume the parameters are always shared between the embedding and output layers to improve sample efficiency.
Since other mobility related attributes such as day of week, time, and duration are considered to be useful for next place prediction, we use them as input in addition to the place itself, by discretizing time and duration and assigning embeddings to these values.


\subsection{Spatial Hierarchical Embeddings}
\label{sec:spatial_embeddings}

\begin{figure}[t]
\centering
  \includegraphics[width=0.85\linewidth]{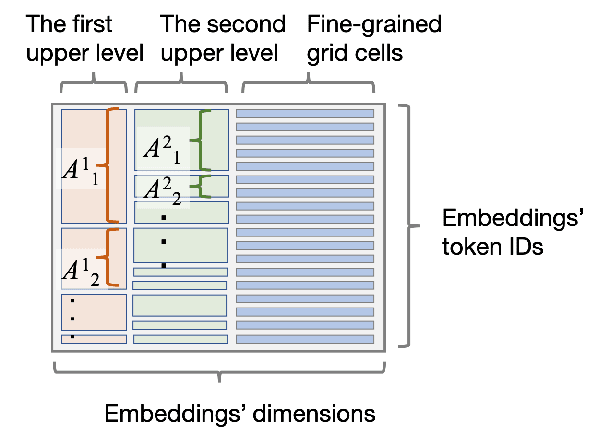}
  \caption{An example layout of the embedding matrix incorporating spatial hierarchy.}
  \label{fig:embd_layout}
\end{figure}

In this section, we describe how we incorporate spatial hierarchy into places and process the embeddings according to the hierarchy in model training.
Consider $L$-levels of spatial hierarchy $\{ 1, \ldots, L \}$ in which each level $l$ is composed of grid cells with sizes of $r^l$ meters.
Among these levels, the first level $1$ has the largest scale, and $r^l$ decreases as $l$ increases.
Those grids are aligned with each other as shown in Fig \ref{fig:hier_embd_overview}.
Now, we represent places contained in an upper level grid cell on a $l$th level as $A^l_k$ using indices $k$ specific to each level.
At the same time, each level has its own $d^l$-dimensional slice within the $d$-dimensional place embedding, leaving a slice with some non-zero dimensionality for the place itself.
In other words, we determine $d^l$ for each $l$ so that the following holds: $\sum_{l=1}^L d^l < d$.
We also denote a slice within an embedding $\mathbf{x}_i$ allocated to a level $l$ by $\mathbf{x}^l_i$.
Furthermore, we refer to a grid cell on a level as a region to clearly distinguish it from the fine grained grid cell.

In this hierarchical structure, all the slices of embeddings belonging to the same upper level region are aggregated so that all of the encompassed lower level regions and places share a common uniform embedding slice corresponding to that upper level region.
We can define this operation of averaging slices and assigning the result back to them as follows: 
\begin{align}
    x^l_i = \frac{\sum_{j \in A^l_k} x^l_j}{ | A^l_k | }, \forall i \in A^l_k,
\end{align}
for the $k$th region on level $l$, and we conduct this for each region on each level $l \in L$.
We repeat this set of averaging operations with short intervals in training, or with no interval between iterations to make slices belonging to an upper region consistently uniform.

In an embedding matrix in general, entries are often sorted in the decreasing order of appearing frequency.
However, reading out from and writing back to vectors in the embedding matrix for our averaging operation takes time proportional to $| A^l_k |$ as slices to be accessed are in most cases scattered in the matrix if they are normally arranged.
To make this operation computationally tractable, we propose a different way of arranging entries in the embedding matrix as shown in Fig \ref{fig:embd_layout} so that slices belonging to the same region on an upper level are grouped together reflecting the structure of spatial hierarchy.
Under this setting, a group of slices can be read and written at once for the averaging operation, at least on the abstraction level of N-dimensional array calculation in high-level programming languages.





\section{Experiments}
\label{sec:exp}
We evaluated our method through the next place prediction task, which is also the modeling task that is used to generate the place embeddings.
The performance of the next place prediction reflects the quality of the obtained place embeddings.

\subsection{Settings for next place prediction task }
\begin{table}[tbp]
  \centering
  \begin{tabular}{lrrrr}
  \toprule
  City & \# of seq. & Vocab. size & Level 1 & Level 2 \\ \midrule
  Fukuoka & 75,363 & 312,112 & 4,467 & 43,981 \\
  Kyoto & 51,463 & 267,881 & 4,910 & 43,219 \\
  Okayama & 46,011 & 217,409 & 3,479 & 32,836 \\ \bottomrule
  \end{tabular}
  \caption{The number of sequences in the training set and the vocabulary size of places and upper-level regions for next place prediction.}
  \label{tbl:npp_datasets}
\end{table}

\begin{figure}[tbp]
  \includegraphics[width=1.0\linewidth]{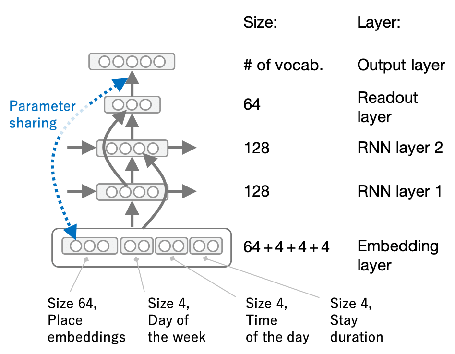}
  \caption{The overview of the LSTM RNN model used in next place prediction.}
  \label{fig:lstm}
\end{figure}

\textbf{Data:}
For each of three cities, we prepared a dataset for embedding generation using human mobility trajectories.
We mapped the sequence of places, timestamps, and duration of stays into token IDs, following the procedure we described in \ref{sec:spatial_embeddings} when issuing token IDs of places.
Also, in this step, we used 10km grid cells as the first upper level and 1km grid cells as the second upper level in the spatial hierarchy and aligned token IDs of places with the structure in the upper levels.
At the same time, we factored each timestamp into day of week and time of day.
Some places in sequences are found to be located outside of the corresponding city, but nevertheless, we include them in the dataset. 
In the data preparation for this training task, we allocated 10\% to the validation and test set respectively, and the remaining 80\% to the training set for each city.
We describe the details of the datasets such as the numbers of records and vocabulary sizes in Table \ref{tbl:npp_datasets}.

\noindent
\newline
\textbf{Model:}
For this task, we applied a 2-layer LSTM RNN \cite{hochreiter97:_long_short_term_memor} with 128-dimensional hidden layers to the human mobility trajectories as illustrated in Fig \ref{fig:lstm}.
We set the size of embeddings to 64 dimensions after confirming that, with this size, the model does not severely underfit but usually mildly overfits during training.
The model has a 64-dimensional readout layer summarizing the information of both two layers of LSTM RNN which then connects to the output layer.
Under this setting, since the input of the output layer and output of the embedding layer are the same size, we have these two layers share the weight parameters except for the biases.
The model was trained over 40 epochs to make sure that the validation performance peaked out during training.
We used Adam \cite{kingma2014adam} for the optimizer, applying clipping norm to prevent the model from breaking down due to exploding gradients.

\noindent
\newline
\textbf{Training:}
For each combination of cities and methods, we repeated the training session 10 times with varying random seed for parameter initialization.
In each session, we used a model with the least validation loss for evaluation with the test set, reporting the average of the test losses over 10 runs as the metrics.
Throughout training sessions, we kept applying the set of averaging operations of slices in embedding vectors explained in \ref{sec:spatial_embeddings} once in 10 iterations. 


\subsection{Comparative Methods}
For each of the three cities, Fukuoka, Kyoto, and Okayama, we evaluated the following four methods.

\noindent
\newline
\textbf{\texttt{hier}}: The first one is our proposed method using two upper levels of the spatial hierarchy, 10km and 1km grid cells, in addition to the most fine grained 125m grid cells.
Among 64 dimensions within the embedding vector, we allocated the first 12 dimensions to the 10km grid cells and the following 20 dimensions to the 1km grid cells. 
The remaining 32 dimensions are used to represent 125m grid cells.
We refer to this proposed method as {\tt hier}.

\noindent
\newline
\textbf{\texttt{hier1km}}: The second one is a variant of the proposed method in which only 1km grid cells were used to represent the upper level in the spatial hierarchy.
We allocated the first 20 dimensions, which is the same size with the span for the 1km grid cells in {\tt hier}, to the upper level and the rest to 125m grid cells.
We refer to this method as {\tt hier1km}.

\noindent
\newline
\textbf{\texttt{hier10km}}: The third method is another variant only using coarser 10km grid cells for the upper level in the hierarchy.
We allocated the first 12 dimensions to the upper level of the spatial hierarchy and the rest to 125m grid cells.
We refer to this method as {\tt hier10km}.

\noindent
\newline
\textbf{\texttt{nonhier}}: The last one is a baseline method in which all the 64 dimensions in the embeddings were used to represent 125m grid cells without considering spatial hierarchy.
We refer to this method as {\tt nonhier}.


\subsection{Next place prediction performance}

\begin{table}[tbp]
  \centering
  \begin{tabular}{llc}
  \toprule
  City & Method & Loss $\pm \, \sigma$ \\ \midrule
  Fukuoka & {\tt hier} & \textbf{7.8406} $\pm$ 0.0097 \\
   & {\tt hier1km} & 7.8975 $\pm$ 0.0140 \\
   & {\tt hier10km} & 7.9641 $\pm$ 0.0184 \\
   & {\tt nonhier} & 8.0523 $\pm$ 0.0127 \\ \midrule
  Kyoto & {\tt hier} & \textbf{8.0546} $\pm$ 0.0202 \\
   & {\tt hier1km} & 8.1342 $\pm$ 0.0139 \\
   & {\tt hier10km} & 8.2058 $\pm$ 0.0118 \\
   & {\tt nonhier} & 8.2785 $\pm$ 0.0134 \\ \midrule
  Okayama & {\tt hier} & \textbf{7.8707} $\pm$ 0.0111 \\
   & {\tt hier1km} & 7.9463 $\pm$ 0.0155 \\
   & {\tt hier10km} & 8.0639 $\pm$ 0.0135 \\
   & {\tt nonhier} & 8.1506 $\pm$ 0.0118 \\ \bottomrule
  \end{tabular}
  \caption{Average loss with the standard deviation for the next place prediction task.}
  \label{tbl:npp_loss}
\end{table}

Table \ref{tbl:npp_loss} shows the experimental results for the next place prediction task.
The numbers represent the test loss, or more specifically, the log-perplexity of the prediction over the test set.
In all of the three cities which we experimented on, our proposed method {\tt hier} gave the best performance.
In addition, the order {\tt hier}, {\tt hier1km}, {\tt hier10km}, and {\tt nonhier} in the performance holds for all the cities.
While the differences among the methods were not large in values themselves, p-values for t-test were smaller than $10^{-6}$ for all the adjoining pairs in the order for each city.
Therefore, the differences in the model performances are confirmed to be statistically significantly.

From these results, we can see that using both 1km and 10km grid cells as upper levels in the spatial hierarchy was useful for next place prediction based on fine grained 125m grid cells.
The spatial level with $\times 8$ larger scale, which were 1km grid cells, contributed more than the other one with $\times 80$ scale, which captures more coarser properties of the places.
Using the combination of the two hierarchical levels surpassed the others even though the number of dimensions allocated to the fine grained 125m grid cells was only half of that of the non-hierarchical baseline method.

\section{Application to Land Use Classification}
To measure how the learned fine grained place embeddings, described in Section \ref{sec:exp}, can represent places accurately, we apply them to a classification task using the Urban Area Land Use Mesh Data.
Land use classification is a popular and complex task in computational urban planning, which usually requires various data sources including satellite imagery, social media, and location data \cite{zhang2019functional}.
To verify the quality of the embeddings, we test whether our fine grained place embeddings can solve the land use classification problem.

\begin{table}[tbp]
  \centering
  \begin{tabular}{llrrr}
  \toprule
  \multirow{2}{*}{acr.} & \multirow{2}{*}{Land use label} & \multicolumn{3}{c}{City} \\ \cmidrule{3-5}
  & & Fukuoka & Kyoto & Okayama \\ \midrule
  far. & Farmland & 215 & 113 & 1,445 \\
  for. & Forest & 434 & 470 & 1,238 \\
  was. & Waste  & 4 & 2 & 7 \\
  hig. & High-rise bldgs & 85 & 39 & 2 \\
  fac. & Factories & 45 & 29 & 169 \\
  low. & Low-rise bldgs & 1,387 & 999 & 1,555 \\
  roa. & Roads & 5 & 3 & 8  \\
  rai. & Railroads & 0 & 5 & 4  \\
  pub. & Public facilities  & 66 & 11 & 55 \\
  vac. & Vacant ground  & 39 & 12 & 47 \\
  par. & Parks & 21 & 0 & 5  \\
  riv. & Rivers and lakes & 19 & 84 & 249 \\
  seas. & Seashore & 4 & 0 & 0  \\
  seaa. & Sea areas & 51 & 0 & 76 \\
  gol. & Golf courses  & 14 & 6 & 13 \\ \midrule
  \multicolumn{2}{c}{(Total)} & 2,389 & 1,773 & 4,873 \\ \bottomrule
  \end{tabular}
  \caption{List of the land use labels and total number of labels for each city.}
  \label{tbl:land_use_labels}
\end{table}

\noindent
\newline
\textbf{Data: } To prepare the data, we aggregated the 17 labels into 15 with similar meanings, as listed in Table \ref{tbl:land_use_labels}.
The learned embeddings correspond to 125m grid cells (one eighth of 1km), but the land use labels are assigned to 100m grid cells.
We addressed this mismatch by aggregating the labels into 500m grid cells, which are aligned with 125m grid cells, using the most frequent label within a cell as a new label.
After that, we prepared classification datasets assigning aggregated labels to individual embeddings and generating one dataset from each model learned by next place prediction.
For each dataset, we allocated 10\% to the validation and test set respectively, and the remaining 80\% to the training set. 
The places used for training, validation, and test were fixed for all 4 methods to enable a fair comparison between the embedding methods. 

\begin{table}[t]
  \centering
  \begin{tabular}{llcc}
  \toprule
  \multirow{2}{*}{City} & \multirow{2}{*}{Method} & \multicolumn{2}{c}{Accuracy $\pm \, \sigma$} \\ \cmidrule{3-4}
   &  & All Places & Rural Places  \\ \midrule
  Fukuoka & {\tt hier} & \textbf{0.725} $\pm$ 0.015  & \textbf{0.601} $\pm$ 0.021 \\
   & {\tt hier1km} & 0.701 $\pm$ 0.022 & 0.565 $\pm$ 0.030 \\
   & {\tt hier10km} & 0.669 $\pm$ 0.019 & 0.505 $\pm$ 0.023 \\
   & {\tt nonhier} & 0.671 $\pm$ 0.010 & 0.495 $\pm$ 0.009 \\ \midrule
  Kyoto & {\tt hier} & 0.739 $\pm$ 0.011 & 0.709 $\pm$ 0.025 \\
   & {\tt hier1km} & \textbf{0.776} $\pm$ 0.004 & \textbf{0.760} $\pm$ 0.011 \\
   & {\tt hier10km} & 0.728 $\pm$ 0.005 & 0.676 $\pm$ 0.011 \\
   & {\tt nonhier} & 0.690 $\pm$ 0.004  & 0.622 $\pm$ 0.012 \\ \midrule
  Okayama & {\tt hier} & \textbf{0.602} $\pm$ 0.005 & \textbf{0.578} $\pm$ 0.008 \\
   & {\tt hier1km} & 0.519 $\pm$ 0.009  & 0.528 $\pm$ 0.018 \\
   & {\tt hier10km} & 0.507 $\pm$ 0.005  & 0.506 $\pm$ 0.008 \\
   & {\tt nonhier} & 0.543 $\pm$ 0.005  & 0.499 $\pm$ 0.010 \\ \bottomrule
  \end{tabular}
  \caption{Average accuracy with the standard deviation for land use classification for all places and only lower 30 percentile places (``rural places'') with regard to the number of visits.}
  \label{tbl:land_use_accuracy}
\end{table}


\begin{figure}[t]
  \centering
  \includegraphics[width=\linewidth]{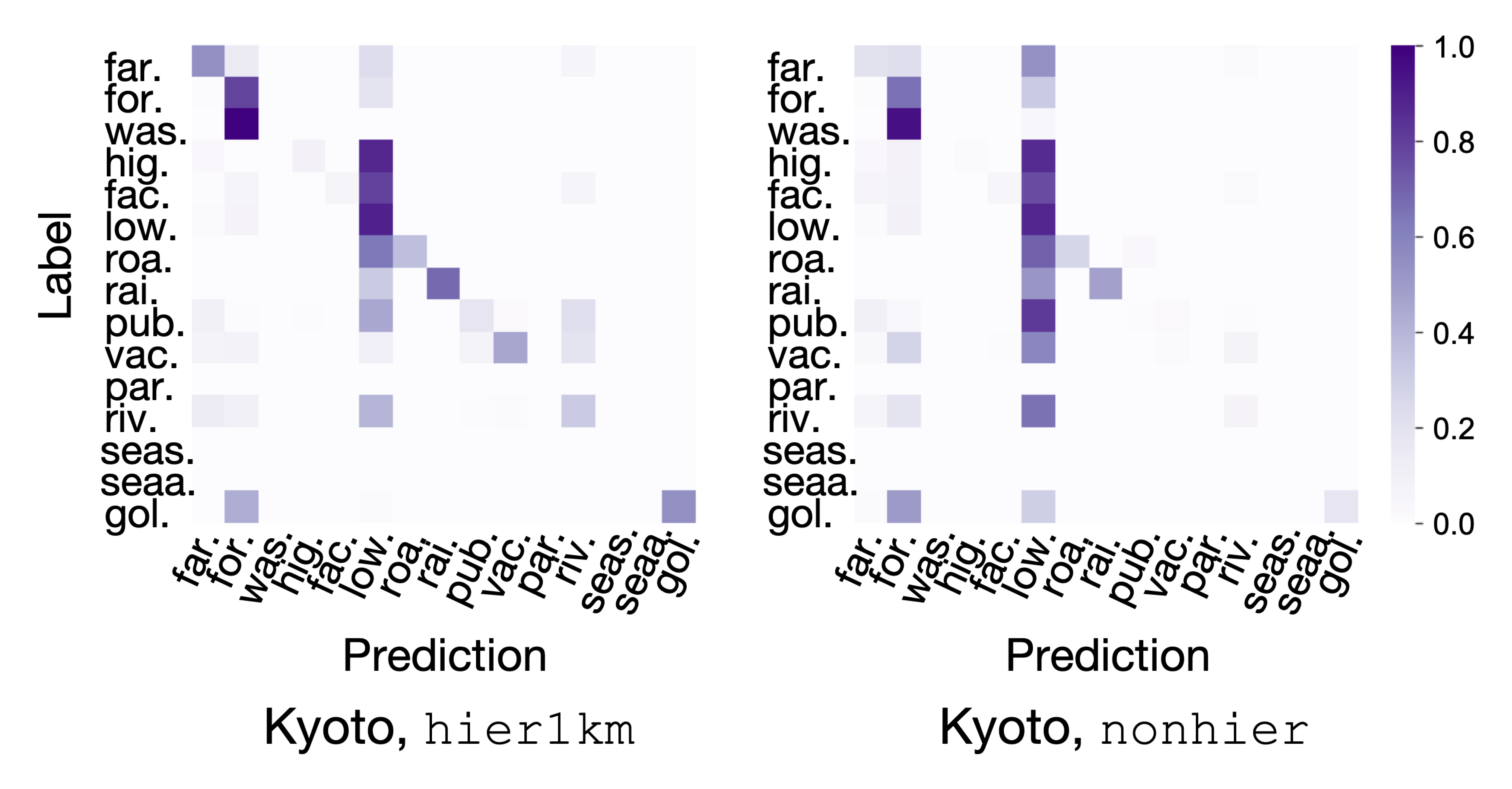}
  \caption{Confusion matrix of the classification result for {\tt hier1km} and {\tt nonhier} for Kyoto city.}
  \label{fig:cm_kyoto-comparison}
\end{figure}

\begin{figure*}[t]
  \centering
  \includegraphics[width=\linewidth]{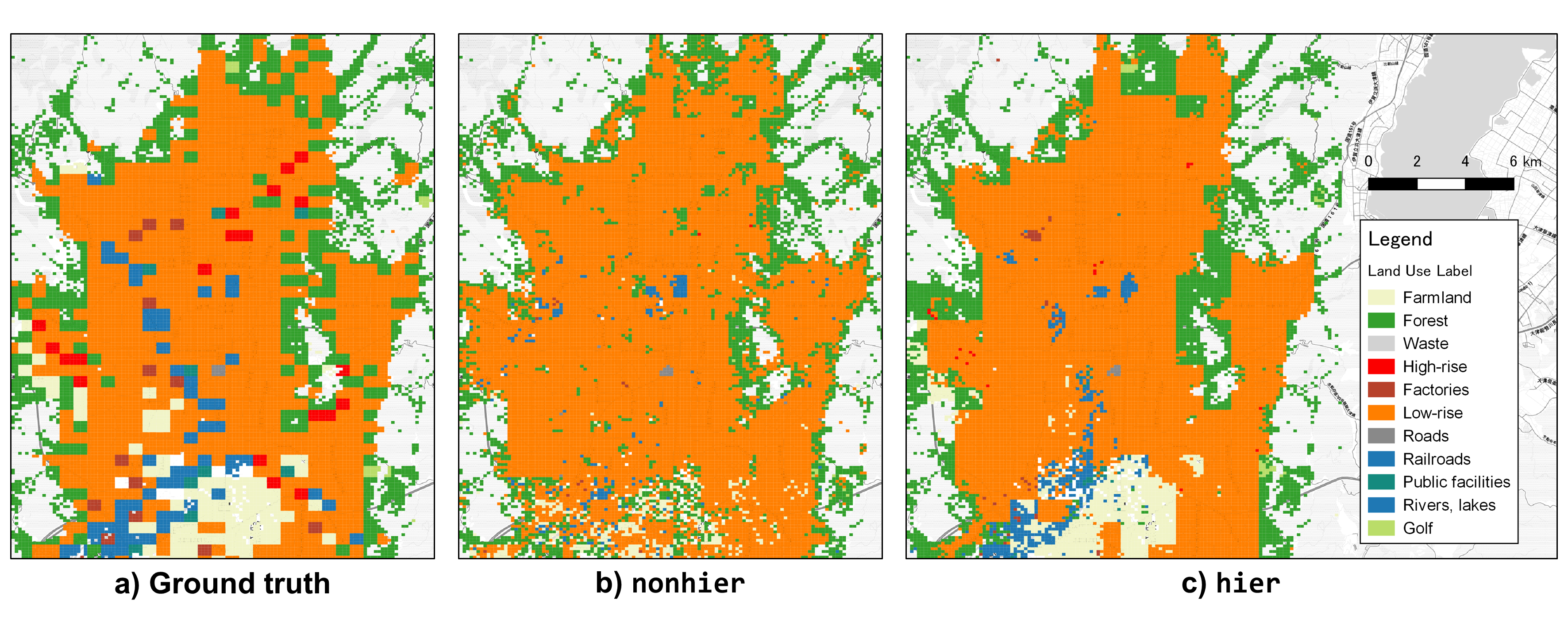}
  \caption{Visualization of estimated land use types in Kyoto City. a) Ground truth, b) \texttt{nonhier}, and c) {\tt hier1km}.}
  \label{fig:landuse}
\end{figure*}

\noindent
\textbf{Model: } To test the predictability of land use labels using the generated place embeddings, we used a simple neural network architecture consisting of one fully-connected layer with a 64 dimensional input for an embedding and a 15 dimensional output for labels so that we can evaluate linear separability of the learned embedding space.
We trained the classification model over 200 epochs using each dataset and evaluated the test accuracy using a model with the best validation accuracy.
The results are shown in Table \ref{tbl:land_use_accuracy}.
We have 10 datasets for each combination of a city and a method, thus taking average over 10 values of the accuracy.


\noindent
\newline
\textbf{Result: }Overall, our main proposed method \texttt{hier} showed the best performance for two cities, as shown in Table \ref{tbl:land_use_accuracy}.
However, for Kyoto, \texttt{hier1km} gave a significantly better accuracy than the others, indicating that global information of a 10km $\times$ 10km area is not informative, if not distracting for this task. 
This suggests that hyperparameter selection is not a trivial task, and that there is room for future research to find optimal choices of the combinations of upper-levels in the spatial hierarchy, and also on the allocation of embeddings' dimensions to each of the levels.
The confusion matrices for the prediction results of models \texttt{hier1km} and \texttt{nonhier} are shown in Figure \ref{fig:cm_kyoto-comparison}. 
We can observe that the fine grained place embeddings with spatial hierarchy (\texttt{hier1km}) perform particularly well for less frequently existing land use types, such as public facilities, vacant ground, parks, rivers and lakes, and golf courses, shown in the bottom right region of the confusion matrices. 
On the other hand, we can observe from the right panel of Figure \ref{fig:cm_kyoto-comparison} that the non-hierarchical embedding incorrectly classifies various land use types including high-rise buildings, factories, rivers and lakes, and golf courses, as low-rise buildings.
To qualitatively assess the performances of the land use classification task, we visualize in Figure \ref{fig:landuse} the a) ground truth labels, predicted land use labels using b) \texttt{nonhier} model, and c) \texttt{hier1km} model for Kyoto City. 
All places in train, validation, and test datasets are included in the visualization. 
By comparing the ground truth and the estimated results, we can visually confirm that the \texttt{hier1km} model is able to capture subtle land use patterns. 
For example, \texttt{hier1km} captures the farmland in the southern areas, the river going through the city from the south, high-rise areas shown in red, and the factory areas shown in brown, where the \texttt{nonhier} labels all of them as either low-rise buildings or forests, which are the major land use types in the city. 

Moreover, to demonstrate the advantage of our model, we also show the prediction accuracy of predicting labels of less-frequently visited places with less data. 
Here, we used the places in the lower 30 percentile in terms of the number of visits (named as ``rural places'') for prediction. 
The right column in Table \ref{tbl:land_use_accuracy} shows the means and standard deviations of accuracy using each model in each city. 
We confirm that overall tendencies about which method performs well and which does not are similar between results for rural areas and all places, the latter being able to make more accurate predictions.

\section{Related Works}
Learning the embeddings of places and locations has been a popular research topic in the urban computing field \cite{zheng2014urban}, which is applied to solve various tasks.
These tasks include next place prediction (e.g. \cite{liu2016predicting}), place-of-interest recommendation (e.g. \cite{chang2018content}) for location service users, and open-site selection for new enterprises (e.g. \cite{xu2016}).
Recent developments in the natural language processing field on learning embeddings of words, such as \texttt{word2vec} \cite{mikolov2013distributed} and \texttt{GloVe} \cite{Pennington14glove:global}, have triggered the recent trend of developing methods for learning effective representations.
Models such as \texttt{POI2vec} have applied ideas similar to word embeddings using social media check-in data, where the analogy of POIs = words, sequences of POIs = sentences was made \cite{liu2016exploring,feng2017poi2vec}. 
A recent model named \texttt{CAPE} used Instagram data as input, where both the location and text data were used to generate embeddings of POIs \cite{chang2018content}. 
A variety of other works have used different types of data, such as flickr \cite{yang2017robust}. 
Models such as \texttt{Geo-Teaser} and \texttt{Place2vec} have combined the users' check-in data and the geographical proximity of POIs to create embeddings \cite{zhao2017geo,yan2017itdl,kong2018hst}. 

With the recent increase in availability of large scale mobility data such as mobile phone location data \cite{gonzalez2008understanding,ratti2006mobile} and taxi GPS data, recent studies have applied the embedding methods to mobility data with more spatio-temporal detail \cite{zhou2018deepmove,wang2017region}.
A recent study created embeddings of GPS coordinates from large scale location datasets \cite{yin2019gps2vec}.
\texttt{ZE-Mob} created embeddings of places using the New York Taxi GPS dataset \cite{yao2018representing}.
A different study proposed a model to create embeddings of human mobility trajectories instead of places \cite{gao2017identifying}.
Although various methods have been proposed to generate place embeddings, all of these methods have the drawback of requiring an abundance of training data in the area of interest.
This motivates us to propose a novel method to overcome data sparsity, which does not necessarily compete against these methods but rather complements the models to produce embeddings robust to data sparsity.

\section{Conclusion}
We proposed a method to produce place embeddings from human mobility trajectory data.
While such place embeddings with high spatial resolution tend to suffer from data sparsity especially in less populated areas, our method can mitigate it by enabling fine grained embeddings to collectively learn places' coarse scale properties, utilizing hierarchical structures in geo-location data.
We evaluated performance of the proposed hierarchical method on next place prediction tasks and found that the proposed hierarchical place embeddings improved prediction accuracy.
Moreover, we tested the learned embeddings' quality through land use classification and clarified that embeddings learned by the proposed method performed better than those of existing embedding methods.
Our proposed method can be applied in a complementary manner to other place embedding methods.


\bibliographystyle{named}
\bibliography{ijcai19}

\end{document}